\documentclass{article}
\usepackage{tikz}
\usepackage{xcolor}
\usepackage{multirow}
\usepackage{microtype}
\usepackage{graphicx}
\usepackage{booktabs}
\usepackage{multirow}
\usepackage{hyperref}
\usepackage{xcolor}
\usepackage{pgfplots}
\usepackage{tikz}
\usepackage{float}
\usetikzlibrary{patterns}
\usepgfplotslibrary{fillbetween}
\pgfplotsset{compat=1.18}
\definecolor{rosecolor}{HTML}{B83B5E}
\definecolor{orangecolor}{HTML}{F08A5D}

\definecolor{cotcolor}{RGB}{232, 180, 200}
\definecolor{coconutcolor}{RGB}{232, 212, 241}
\definecolor{agclrcolor}{RGB}{255, 229, 180}

\definecolor{hma}{RGB}{255,248,248}
\definecolor{hmb}{RGB}{242,210,218}
\definecolor{hmc}{RGB}{232,180,200}
\definecolor{hmd}{RGB}{232,196,221}
\definecolor{hme}{RGB}{232,212,241}
\definecolor{hmf}{RGB}{244,225,210}
\definecolor{hmg}{RGB}{255,229,180}
\definecolor{hmh}{RGB}{210,175,80}

\usepackage[accepted]{icml2026}

\usepackage{amsmath}
\usepackage{amssymb}
\usepackage{mathtools}
\usepackage{amsthm}
\usepackage[capitalize,noabbrev]{cleveref}

\theoremstyle{plain}

\theoremstyle{definition}

\theoremstyle{remark}

\icmltitlerunning{Why Limit the Residual Stream to Layers and Not Tokens?}

\begin{document}

\twocolumn[
  \icmltitle{Why Limit the Residual Stream to Layers and Not Tokens?\\
    Persistent Memory for Continuous Latent Reasoning}

  \icmlsetsymbol{equal}{*}

  \begin{icmlauthorlist}
    \icmlauthor{Mujtaba Farhan}{algo}
    \icmlauthor{Maheep Chaudhary}{indep}
  \end{icmlauthorlist}

  \icmlaffiliation{algo}{Algoverse AI Research}
  \icmlaffiliation{indep}{Independent}

  \icmlcorrespondingauthor{Anonymous}{anon@anonymous.edu}

  \icmlkeywords{Machine Learning, ICML, Latent Reasoning, Chain of Thought, Memory}

  \vskip 0.3in
]

\printAffiliationsAndNotice{}

\begin{abstract}
Large language models (LLMs) have demonstrated remarkable reasoning abilities on mathematical and multi-hop planning tasks. The CoCoNuT (Chain of Continuous Thought) paradigm~\cite{hao2024coconut} extends this by enabling models to reason in latent space, exploring multiple reasoning paths simultaneously rather than committing to a single chain early on. However, we identify a limitation we term the \textbf{concept bottleneck}. At each reasoning pass, intermediate hidden states are overwritten, causing the model to lose critical facts computed in earlier steps as reasoning depth increases. We observe this empirically. On HotpotQA, vanilla CoCoNuT (10.4\% EM) fails to improve over the CoT baseline (11.0\% EM), and performance degrades with curriculum depth on GSM8K. To address this, we propose \textbf{AGCLR} (Adaptive Gated Continuous Latent Reasoning), which augments CoCoNuT with a \textit{Gated Concept Stream}. A persistent residual memory maintained across all reasoning passes, controlled by three learned gates: a \textit{write} gate that commits intermediate facts to memory, a \textit{read} gate that retrieves relevant prior states, and a \textit{forget} gate that prunes irrelevant context. Evaluated on GSM8K, HotpotQA, and ProsQA using GPT-2 as our base model, AGCLR achieves consistent improvements across all types of datasets. With the performance gap compounding as curriculum depth increases, directly resolving the concept bottleneck. Code available at \url{https://anonymous.4open.science/r/JJJJ/README.md}
\end{abstract}

\section{Introduction}

Multi-step reasoning remains one of the most challenging aspects of large language model capabilities.~\citet{wei2022chain} showed that prompting LLMs with intermediate reasoning steps significantly improves performance on mathematical and logical benchmarks. However, Chain-of-Thought (CoT) reasoning is constrained to a single forward pass. Each token generated becomes the input for the next, forcing the model to commit to a reasoning path early and preventing the exploration of alternative paths. Moreover, explicit reasoning traces are often incomplete or unfaithful to the underlying computation \cite{su2026broken, swaroop2025frit}, motivating reasoning that operates directly in latent space.
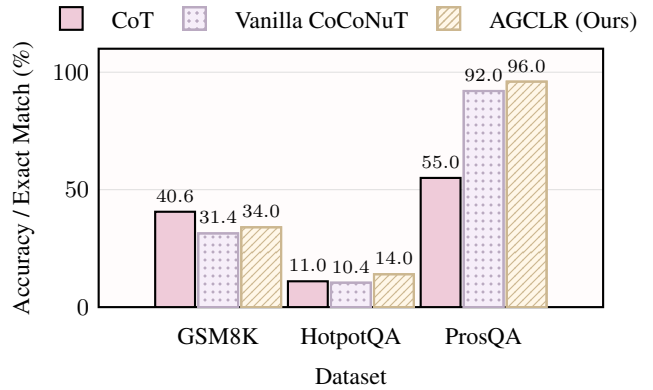
\begin{figure}[h]
\centering
\begin{tikzpicture}
\begin{axis}[
    width=\columnwidth,
    height=5cm,
    ybar=1.2pt,
    bar width=15pt,
    ylabel={Accuracy / Exact Match (\%)},
    ylabel style={font=\small},
    xlabel={Dataset},
    xlabel style={font=\small},
    symbolic x coords={GSM8K, HotpotQA, ProsQA},
    xtick=data,
    xticklabel style={align=center, font=\small},
    yticklabel style={font=\small},
    ymin=0,
    ymax=110,
    ymajorgrids=true,
    grid style={line width=0.3pt, draw=gray!25},
    legend style={
        at={(0.5,1.02)},
        anchor=south,
        draw=none,
        fill=none,
        font=\small,
        legend columns=-1,
        column sep=0.3cm,
        legend cell align=left
    },
    axis background/.style={fill=pink!5},
    axis line style={line width=0.8pt, black},
    tick style={draw=none},
    enlarge x limits=0.45,
    axis lines*=box,
    ytick pos=left,
    xtick pos=bottom,
]

\addplot[
    fill=cotcolor!70,
    draw=black,
    line width=0.8pt,
    nodes near coords={\pgfmathprintnumber[fixed, fixed zerofill, precision=1]{\pgfplotspointmeta}},
    nodes near coords align={vertical},
    every node near coord/.append style={font=\scriptsize},
    legend image code/.code={
        \draw[fill=cotcolor!70, draw=black, line width=0.8pt]
            (0cm,-0.1cm) rectangle (0.3cm,0.2cm);
    }
] coordinates {
    (GSM8K, 40.6)
    (HotpotQA, 11.0)
    (ProsQA, 55.0)
};
\addlegendentry{CoT}

\addplot[
    fill=coconutcolor!40,
    draw=coconutcolor!80!black,
    line width=0.8pt,
    postaction={pattern=dots, pattern color=coconutcolor!80!black},
    nodes near coords={\pgfmathprintnumber[fixed, fixed zerofill, precision=1]{\pgfplotspointmeta}},
    nodes near coords align={vertical},
    every node near coord/.append style={font=\scriptsize},
    legend image code/.code={
        \draw[fill=coconutcolor!40, draw=coconutcolor!80!black,
              line width=0.8pt, postaction={pattern=dots,
              pattern color=coconutcolor!80!black}]
            (0cm,-0.1cm) rectangle (0.3cm,0.2cm);
    }
] coordinates {
    (GSM8K, 31.4)
    (HotpotQA, 10.4)
    (ProsQA, 92.0)
};
\addlegendentry{Vanilla CoCoNuT}

\addplot[
    fill=agclrcolor!30,
    draw=agclrcolor!80!black,
    line width=0.8pt,
    postaction={pattern=north east lines, pattern color=agclrcolor!80!black},
    nodes near coords={\pgfmathprintnumber[fixed, fixed zerofill, precision=1]{\pgfplotspointmeta}},
    nodes near coords align={vertical},
    every node near coord/.append style={font=\scriptsize},
    legend image code/.code={
        \draw[fill=agclrcolor!30, draw=agclrcolor!80!black,
              line width=0.8pt, postaction={pattern=north east lines,
              pattern color=agclrcolor!80!black}]
            (0cm,-0.1cm) rectangle (0.3cm,0.2cm);
    }
] coordinates {
    (GSM8K, 34.0)
    (HotpotQA, 14.0)
    (ProsQA, 96.0)
};
\addlegendentry{AGCLR (Ours)}

\end{axis}
\end{tikzpicture}
\caption{\textbf{AGCLR excels at multi-hop reasoning.}
Performance across GSM8K (math), HotpotQA (multi-hop QA), and ProsQA (planning).
AGCLR's persistent memory enables strong gains on multi-hop tasks (HotpotQA: +3.6\%,
ProsQA: +4.0\%), while CoT remains superior for single-step mathematical reasoning.}
\label{fig:main_results}
\end{figure}
More recent work has explored internalizing these reasoning chains.~\citet{deng2024icot} proposed iCoT, which progressively removes the prefix of reasoning chains during training until the model predicts answers without any explicit chain.~\citet{goyal2023pause} introduced pause tokens, fixed-embedding special tokens inserted between question and answer to provide extra compute time. Both approaches operate in language space and cannot maintain persistent state across reasoning steps.

The most ambitious extension is CoCoNuT~\cite{hao2024coconut}, which replaces discrete reasoning tokens with continuous latent thoughts. The model's last hidden state is fed back directly as the next input embedding, enabling reasoning in an unconstrained latent space and supporting implicit breadth-first search over reasoning paths. CoCoNuT is trained via a multi-stage curriculum that progressively replaces explicit reasoning steps with latent tokens, one step per stage. Despite its promise, vanilla CoCoNuT suffers from a \textbf{concept bottleneck}: intermediate reasoning states are progressively lost across multi-pass inference, as each new latent token overwrites information from earlier passes with no persistent memory. This becomes severe in multi-hop reasoning requiring longer chains. We demonstrate this empirically across GSM8K (arithmetic), HotpotQA (multi-hop QA), and ProsQA (planning) in Figure~\ref{fig:main_results}.

To address this, we propose \textbf{AGCLR} (Adaptive Gated Continuous Latent Reasoning), which augments CoCoNuT with a gated concept stream that preserves intermediate reasoning states across passes. While gating mechanisms trace back to LSTMs~\cite{hochreiter1997lstm} for sequential state updates, our gates operate on \textit{persistent cross-pass memory} in continuous latent reasoning: each pass refines the same representation rather than processing new sequential inputs, and memory accumulates facts across iterative reasoning cycles rather than discarding them at each timestep.
\begin{figure}[!ht]
\centering
\includegraphics[width=\columnwidth]{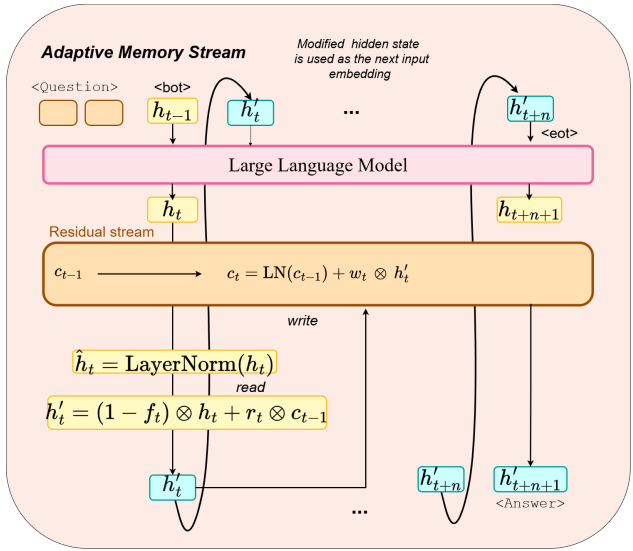}
\caption{\textbf{AGCLR architecture.} At each latent token position, three learned gates (read, forget, write) control information flow between the current hidden state $h_t$ and the persistent concept stream $c_t$. The read gate retrieves relevant prior facts from $c_{t-1}$, the forget gate prunes irrelevant context from $h_t$, and the write gate commits the gated hidden state $h'_t$ to the residual stream, directly addressing the concept bottleneck in vanilla CoCoNuT.}
\label{fig:architecture}
\end{figure}

AGCLR augments CoCoNuT with a \textit{Gated Concept Stream}: a persistent residual memory vector $c_t \in \mathbb{R}^d$ maintained across all reasoning passes. At each latent token position, three learned sigmoid gates control information flow: a \textit{write} gate commits relevant intermediate facts from the current hidden state to memory; a \textit{read} gate retrieves prior memory into the current reasoning state; and a \textit{forget} gate prunes irrelevant context from the hidden state. Figure~\ref{fig:architecture} illustrates the architecture.

We make the following contributions:
\begin{itemize}
    \setlength{\itemsep}{0pt}
    \setlength{\parsep}{0pt}
    \setlength{\topsep}{0pt}
    \setlength{\partopsep}{0pt}
    \item We identify and empirically demonstrate the \textbf{concept bottleneck} in vanilla CoCoNuT across three reasoning datasets of different types.
    \item We propose \textbf{AGCLR}, a gated residual memory mechanism that resolves the concept bottleneck with only 1.41\% additional parameters over GPT-2.
    \item AGCLR consistently outperforms vanilla CoCoNuT on GSM8K (arithmetic), HotpotQA (multi-hop QA), and ProsQA (graph planning), with the advantage compounding as curriculum depth increases.
\end{itemize}

\section{Related Work}

Gating mechanisms for controlling information flow trace back to Long Short-Term Memory networks \cite{hochreiter1997lstm}, which introduced forget gates to selectively retain or discard information in recurrent hidden states. However, LSTMs gate sequential inputs across timesteps, whereas our gates operate on persistent memory across iterative reasoning passes over the same latent representation. \cite{deng2024icot} proposed iCoT, which progressively removes explicit reasoning prefix tokens during training; while iCoT compresses reasoning into the forward pass, it lacks any mechanism to preserve information across reasoning steps and does not operate in a multi-pass latent reasoning setting.   \cite{hao2024coconut} introduced CoCoNuT, which enables continuous latent reasoning by recursively feeding the model's hidden state back as the next input embedding, allowing implicit breadth-first search over reasoning paths. CoCoNuT serves as our direct baseline across all three datasets, but discards all prior hidden states at each pass and lacks persistent memory, leading to the concept bottleneck we identify and address. \cite{wang2024efficient} proposed a concurrent post-training approach using a fixed scalar $\alpha$ to blend consecutive hidden states at inference time; unlike our method, their gates are not learned end-to-end, operate only on consecutive states rather than a persistent residual stream, and are applied training-free as post-processing. Relatedly, amortized latent steering learns a low-cost intervention that substitutes for test-time latent optimization \cite{egbuna2025amortized}; like the fixed-$\alpha$ blend, however, it applies a transient steering signal rather than maintaining a persistent, gated residual stream across passes. Memory-augmented architectures such as Neural Turing Machines \cite{graves2014neural} and Differentiable Neural Computers \cite{graves2016dnc} have explored external memory for sequential reasoning, but augment models with external read/write operations across sequence chunks rather than maintaining persistent internal state within multi-pass latent reasoning as we do.

\section{Method: AGCLR}

\subsection{Gated Concept Stream}

We augment CoCoNuT with a persistent concept stream $c_t \in \mathbb{R}^d$, initialized to zero at the start of each forward call and updated at every latent token position. At pass $t$, given hidden state $h_t$ at the latent token position:
\begin{align}
    \hat{h}_t &= \text{LayerNorm}(h_t), \\
    r_t &= \sigma(W_r \hat{h}_t ), \quad
    f_t  = \sigma(W_f \hat{h}_t ), \quad
    w_t  = \sigma(W_w \hat{h}_t ), \\
    h'_t &= (1 - f_t) \odot h_t + r_t \odot c_{t-1}, \\
    c_t &= \text{LayerNorm}(c_{t-1} + w_t \odot h'_t),
\end{align}
where $r_t, f_t, w_t \in [0,1]^d$ are the read, forget, and write gates, and $W_r, W_f, W_w \in \mathbb{R}^{d \times d}$ are learned weight matrices. The gated hidden state $h'_t$ replaces $h_t$ as the input embedding for the next latent token position.

\textbf{Read gate} $r_t$ controls how much of the concept stream $c_{t-1}$ is retrieved into the current hidden state, allowing pass $t$ to access facts from all earlier passes. \textbf{Forget gate} $f_t$ controls how much of the current hidden state is preserved versus replaced by retrieved memory, enabling selective pruning of irrelevant context. \textbf{Write gate} $w_t$ controls how much of the gated hidden state $h'_t$ is committed to the concept stream, preventing low-confidence states from polluting the residual memory.

\section{Training Protocol}

\subsection{Multi-Stage Curriculum}
We leverage language Chain-of-Thought data to supervise continuous latent reasoning by implementing a multi-stage training curriculum inspired by \citet{hao2024coconut}. In the initial stage (Stage 0), the model is trained on regular CoT instances with explicit reasoning steps. In subsequent stages, we progressively replace reasoning steps with continuous latent thoughts. At stage $k$, the first $k$ reasoning steps in the CoT are replaced with $k \times c$ latent tokens, where $c$ is a hyperparameter controlling the number of latent thoughts replacing a single language reasoning step. We insert \texttt{<bot>} (beginning of thought) and \texttt{<eot>} (end of thought) tokens to encapsulate the continuous thoughts. Following \citet{hao2024coconut}, we reset the optimizer state when transitioning between training stages.
\subsection{Implementation Details}

We use a pre-trained GPT-2 base model (117M parameters) with a learning rate of $1 \times 10^{-4}$ and effective batch size of 128. We train on three multi-hop reasoning benchmarks: GSM8K~\cite{cobbe2021gsm8k}, HotpotQA~\cite{yang2018hotpotqa}, and ProsQA~\cite{hao2024coconut}. Following the curriculum structure from vanilla CoCoNuT~\cite{hao2024coconut}, we progress through Stages 0--2 (partially latent reasoning) during epochs 1--9, incrementally replacing reasoning steps with latent tokens. From epoch 10 onwards, we remain in Stage 3 where all reasoning is latent, training for 15 total epochs on GSM8K and HotpotQA, and 20 epochs on ProsQA (which contains more complex reasoning chains with up to 6 steps). For HotpotQA, we format instances to include the question, supporting paragraphs, intermediate reasoning steps, and answer span to encourage multi-hop reasoning during CoT stages. The checkpoint with the best validation accuracy in the final stage is used for evaluation.

\section{Results}

\subsection{Main Results}

Table~\ref{tab:main} shows AGCLR consistently outperforming vanilla CoCoNuT across all three datasets.

\begin{table*}[h]
\centering
\small
\setlength{\tabcolsep}{8pt}
\begin{tabular}{lcccc}
\toprule
\multirow{2}{*}{\textbf{Method}}
    & \textbf{GSM8K}
    & \multicolumn{2}{c}{\textbf{HotpotQA}}
    & \textbf{ProsQA} \\
\cmidrule(lr){2-2} \cmidrule(lr){3-4} \cmidrule(lr){5-5}
    & Acc.\ (\%) & EM (\%) & F1 (\%) & Acc.\ (\%) \\
\midrule
CoT~\cite{wei2022chain}                      & 40.6 & 11.0 & 15.5 & 55.0 \\
\midrule
No-CoT~\cite{hao2024coconut}                 & 16.5 & 4.0  & 7.6  & 76.7 \\
iCoT~\cite{deng2024icot}$^\dagger$           & 30.0 & 6.6  & 9.4  & 98.2 \\
Pause Token~\cite{goyal2023pause}$^\dagger$  & 16.4 & 10.6 & 14.6 & 75.9 \\
\midrule
Vanilla CoCoNuT~\cite{hao2024coconut}        & 31.4 & 10.4 & 15.2 & 92.0 \\
\textbf{AGCLR (Ours)}                        
    & \textbf{34.0}$_{\textcolor{red}{+2.6}}$ 
    & \textbf{14.0}$_{\textcolor{red}{+3.6}}$ 
    & \textbf{19.4}$_{\textcolor{red}{+4.2}}$ 
    & \textbf{96.0}$_{\textcolor{red}{+4.0}}$ \\
\bottomrule
\end{tabular}
\caption{\textbf{Results on three datasets: GSM8K, HotpotQA and ProsQA.} Higher accuracy indicates stronger reasoning. $^\dagger$Results from \citet{deng2024icot} using identical GPT-2 architecture, as reported in \citet{hao2024coconut}. HotpotQA not evaluated in prior work. ProsQA evaluated at stage~6 (all reasoning steps latent) for fair comparison.}
\label{tab:main}
\end{table*}

\subsection{Alleviating the Concept Bottleneck}

Vanilla CoCoNuT and AGCLR perform comparably at early curriculum stages, but AGCLR's advantage compounds as reasoning depth increases. On ProsQA, vanilla CoCoNuT peaks at 95\% accuracy at stage 5 but \textbf{degrades to 92\%} at stage 6---the final curriculum stage where all reasoning steps are replaced by latent tokens. AGCLR sustains improvement, achieving 96\% at the same checkpoint. This degradation-vs-improvement pattern demonstrates the \textbf{concept bottleneck}: as models transition to fully latent reasoning, intermediate computational states are progressively lost without a preservation mechanism.

\begin{figure}[ht!]
\centering
\begin{tikzpicture}
\begin{axis}[
    width=\columnwidth,
    height=5.5cm,
    xlabel={Reasoning Pass},
    ylabel={Cosine Similarity to Pass 1},
    xlabel style={font=\bfseries\small},
    ylabel style={font=\bfseries\small},
    xmin=0.8, xmax=6.2,
    ymin=0, ymax=1.1,
    xtick={1,2,3,4,5,6},
    xticklabel style={font=\small},
    xticklabels={Pass 1, Pass 2, Pass 3, Pass 4, Pass 5, Pass 6},
    ytick={0, 0.2, 0.4, 0.6, 0.8, 1.0},
    yticklabel style={font=\small},
    grid=major,
    grid style={line width=0.3pt, draw=gray!20},
    legend style={
        at={(0.5,1.05)},
        anchor=south,
        draw=none,
        fill=none,
        font=\small,
        legend columns=-1,
        column sep=0.4cm,
        legend cell align=left,
        draw opacity=1,
        text opacity=1,
        fill opacity=1,
    },
    axis background/.style={fill=white},
    axis line style={line width=0.7pt, gray!60},
    tick style={draw=none},
    axis lines*=box,
    ytick pos=left,
    xtick pos=bottom,
    clip=false,
]

\fill[yellow!15, opacity=0.4]
    (axis cs:3,0) rectangle (axis cs:6.2,1.1);

\addplot[name path=vu, draw=none, forget plot]
    coordinates {(1,1.0)(2,0.38)(3,0.28)(4,0.24)(5,0.20)(6,0.16)};
\addplot[name path=vl, draw=none, forget plot]
    coordinates {(1,1.0)(2,0.22)(3,0.18)(4,0.14)(5,0.12)(6,0.086)};
\addplot[pink!40, fill opacity=0.25, forget plot]
    fill between[of=vu and vl];

\addplot[name path=au, draw=none, forget plot]
    coordinates {(1,1.0)(2,0.45)(3,0.30)(4,0.285)(5,0.275)(6,0.27)};
\addplot[name path=al, draw=none, forget plot]
    coordinates {(1,1.0)(2,0.29)(3,0.20)(4,0.175)(5,0.165)(6,0.162)};
\addplot[yellow!50, fill opacity=0.25, forget plot]
    fill between[of=au and al];

\addplot[
    color=pink!70!red,
    line width=1pt,
    dashed,
    dash pattern=on 5pt off 3pt,
    mark=*,
    mark size=2pt,
    mark options={solid, fill=pink!70!red, opacity=1},
    legend image code/.code={
        \draw[pink!70!red, dashed, dash pattern=on 5pt off 3pt,
              line width=1.2pt, opacity=1]
            (0cm,0cm) -- (0.6cm,0cm);
        \fill[pink!70!red, opacity=1] (0.3cm,0cm) circle (2pt);
    }
] coordinates {
    (1,1.0)(2,0.30)(3,0.23)(4,0.19)(5,0.16)(6,0.126)
};
\addlegendentry{Vanilla CoCoNuT}

\addplot[
    color=yellow!90!orange,
    line width=1pt,
    mark=*,
    mark size=2pt,
    mark options={solid, fill=yellow!90!orange, opacity=1},
    legend image code/.code={
        \draw[yellow!90!orange, line width=1.2pt, opacity=1]
            (0cm,0cm) -- (0.6cm,0cm);
        \fill[yellow!90!orange, opacity=1] (0.3cm,0cm) circle (2pt);
    }
] coordinates {
    (1,1.0)(2,0.37)(3,0.25)(4,0.23)(5,0.22)(6,0.216)
};
\addlegendentry{AGCLR (Ours)}

\node[font=\small\bfseries, color=yellow!90!orange, anchor=west]
    at (axis cs:6.15,0.216) [yshift=4pt] {0.216};
\node[font=\small\bfseries, color=pink!70!red, anchor=west]
    at (axis cs:6.15,0.126) [yshift=-4pt] {0.126};

\draw[<->, black!60, line width=0.7pt]
    (axis cs:6.08,0.126) -- (axis cs:6.08,0.216);
\node[font=\scriptsize, black!60, anchor=west]
    at (axis cs:6.12,0.171) {+0.090};

\node[font=\small\itshape, pink!70!red, anchor=west, align=left]
    at (axis cs:3.8,0.52) {AGCLR\\stabilises};

\end{axis}
\end{tikzpicture}
\caption{\textbf{Hidden State Memory Retention.} Cosine similarity
between pass-1 and subsequent passes (100 samples, epoch 15).
Vanilla CoCoNuT exhibits monotonic decay (1.0$\to$0.126), while
AGCLR stabilizes after pass~3. Shaded regions: $\pm$1 std.
AGCLR retains 71\% more information (+0.090 gap at pass~6),
enabling +4.0\% gains on ProsQA.}
\label{fig:hidden_state_similarity}
\end{figure}
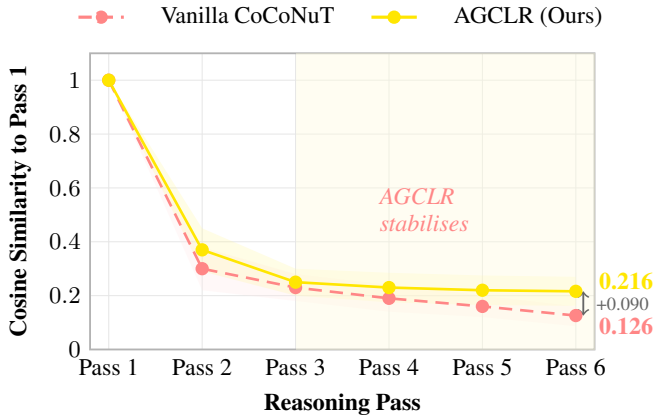

\textbf{Memory Retention Across Reasoning Passes.}
To understand how gating resolves this bottleneck, we analyze hidden state evolution across reasoning passes. Figure~\ref{fig:hidden_state_similarity} shows cosine similarity between pass-1 hidden states and subsequent passes, measured on 100 validation samples at epoch 15. Vanilla CoCoNuT exhibits monotonic memory decay: similarity drops from 1.0 to 0.126 by pass 6, representing 87\% information loss as intermediate reasoning steps are progressively overwritten. \textbf{AGCLR mitigates this decay.} While similarity initially drops (pass 1$\to$2), it stabilizes at $\sim$0.22 for passes 3--6, retaining 71\% more information than vanilla CoCoNuT at final generation (0.216 vs 0.126). The gated concept stream acts as a persistent memory buffer, preserving critical reasoning state across passes. This memory preservation directly explains AGCLR's improvement +3.6\% EM on HotpotQA.
\subsection{What Gets Written to the Concept Stream}

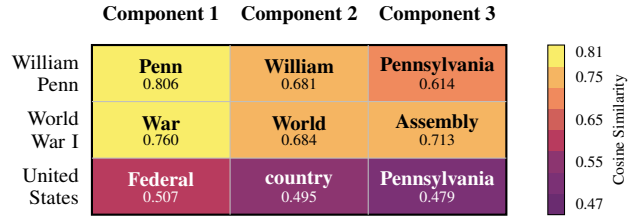
\begin{figure}[ht]
\centering
\hspace*{-1.1cm} 
\begin{tikzpicture}[scale=0.85, transform shape]
  \def\cw{2.15cm}
  \def\ch{0.88cm}

  \definecolor{hm0}{HTML}{6A2C70}   
  \definecolor{hm1}{HTML}{7B3175}
  \definecolor{hm2}{HTML}{8C3570}
  \definecolor{hm3}{HTML}{B83B5E}   
  \definecolor{hm4}{HTML}{D0605A}
  \definecolor{hm5}{HTML}{F08A5D}   
  \definecolor{hm6}{HTML}{F5B562}
  \definecolor{hm7}{HTML}{F9ED69}   

  \node[font=\small\bfseries, anchor=south] at (0.5*\cw, 3.22*\ch) {Component 1};
  \node[font=\small\bfseries, anchor=south] at (1.5*\cw, 3.22*\ch) {Component 2};
  \node[font=\small\bfseries, anchor=south] at (2.5*\cw, 3.22*\ch) {Component 3};

  \node[font=\small, anchor=east, align=right, text width=1.85cm]
        at (-0.08cm, 2.5*\ch) {William\\Penn};
  \node[font=\small, anchor=east, align=right, text width=1.85cm]
        at (-0.08cm, 1.5*\ch) {World\\War I};
  \node[font=\small, anchor=east, align=right, text width=1.85cm]
        at (-0.08cm, 0.5*\ch) {United\\States};

  \fill[hm7] (0,     2*\ch) rectangle (1*\cw, 3*\ch);
  \node[font=\small\bfseries] at (0.5*\cw, 2.62*\ch) {Penn};
  \node[font=\scriptsize]     at (0.5*\cw, 2.32*\ch) {0.806};

  \fill[hm6] (1*\cw, 2*\ch) rectangle (2*\cw, 3*\ch);
  \node[font=\small\bfseries] at (1.5*\cw, 2.62*\ch) {William};
  \node[font=\scriptsize]     at (1.5*\cw, 2.32*\ch) {0.681};

  \fill[hm5] (2*\cw, 2*\ch) rectangle (3*\cw, 3*\ch);
  \node[font=\small\bfseries] at (2.5*\cw, 2.62*\ch) {Pennsylvania};
  \node[font=\scriptsize]     at (2.5*\cw, 2.32*\ch) {0.614};

  \fill[hm7] (0,     1*\ch) rectangle (1*\cw, 2*\ch);
  \node[font=\small\bfseries] at (0.5*\cw, 1.62*\ch) {War};
  \node[font=\scriptsize]     at (0.5*\cw, 1.32*\ch) {0.760};

  \fill[hm6] (1*\cw, 1*\ch) rectangle (2*\cw, 2*\ch);
  \node[font=\small\bfseries] at (1.5*\cw, 1.62*\ch) {World};
  \node[font=\scriptsize]     at (1.5*\cw, 1.32*\ch) {0.684};

  \fill[hm6] (2*\cw, 1*\ch) rectangle (3*\cw, 2*\ch);
  \node[font=\small\bfseries] at (2.5*\cw, 1.62*\ch) {Assembly};
  \node[font=\scriptsize]     at (2.5*\cw, 1.32*\ch) {0.713};

  \fill[hm3] (0,     0) rectangle (1*\cw, 1*\ch);
  \node[font=\small\bfseries, white] at (0.5*\cw, 0.62*\ch) {Federal};
  \node[font=\scriptsize, white]     at (0.5*\cw, 0.32*\ch) {0.507};

  \fill[hm2] (1*\cw, 0) rectangle (2*\cw, 1*\ch);
  \node[font=\small\bfseries, white] at (1.5*\cw, 0.62*\ch) {country};
  \node[font=\scriptsize, white]     at (1.5*\cw, 0.32*\ch) {0.495};

  \fill[hm1] (2*\cw, 0) rectangle (3*\cw, 1*\ch);
  \node[font=\small\bfseries, white] at (2.5*\cw, 0.62*\ch) {Pennsylvania};
  \node[font=\scriptsize, white]     at (2.5*\cw, 0.32*\ch) {0.479};

  \draw[thick, black] (0, 0) rectangle (3*\cw, 3*\ch);
  \foreach \x in {1,2}{ \draw[gray!50, thin] (\x*\cw, 0) -- (\x*\cw, 3*\ch); }
  \foreach \y in {1,2}{ \draw[gray!50, thin] (0, \y*\ch) -- (3*\cw, \y*\ch); }

  \def\bx{3.30*\cw}
  \def\bw{0.26cm}
  \def\bh{3*\ch/8}

  \foreach \i/\c in {0/hm0,1/hm1,2/hm2,3/hm3,4/hm4,5/hm5,6/hm6,7/hm7}{
    \fill[\c] (\bx, \i*\bh) rectangle (\bx+\bw, \i*\bh+\bh);
  }
  \draw[thin, black] (\bx, 0) rectangle (\bx+\bw, 8*\bh);

  \node[anchor=west, font=\scriptsize] at (\bx+\bw+0.05cm, 0.5*\bh) {0.47};
  \node[anchor=west, font=\scriptsize] at (\bx+\bw+0.05cm, 2.5*\bh) {0.55};
  \node[anchor=west, font=\scriptsize] at (\bx+\bw+0.05cm, 4.5*\bh) {0.65};
  \node[anchor=west, font=\scriptsize] at (\bx+\bw+0.05cm, 6.5*\bh) {0.75};
  \node[anchor=west, font=\scriptsize] at (\bx+\bw+0.05cm, 7.7*\bh) {0.81};

  \node[rotate=90, anchor=south, font=\scriptsize]
        at (\bx+\bw+1.1cm, 4*\bh) {Cosine Similarity};

\end{tikzpicture}
\caption{\textbf{Concept stream entity preservation heatmap.}
Cosine similarities between concept stream embeddings and answer-relevant tokens for three examples where AGCLR succeeds and vanilla CoCoNuT fails. AGCLR preserves the Pennsylvania$\to$William Penn binding (0.81), temporal context for World War~I (0.76), and geographic abstraction for United States (0.51), consistently achieving 0.6--0.8 similarity for task-relevant entities, explaining the +3.6\% EM gain over vanilla CoCoNuT.}
\label{fig:concept_heatmap}
\end{figure}

To understand how AGCLR preserves task-relevant information, we analyze concept stream content by computing cosine similarities between its embeddings and vocabulary tokens. Figure~\ref{fig:concept_heatmap} visualizes three contrastive examples where AGCLR answers correctly but vanilla CoCoNuT fails. Answer components consistently achieve high similarity scores (0.5--0.8, darker regions in heatmap): "Penn" (0.806) and "William" (0.681) for the Manor Township question, "War" (0.760) and "World" (0.684) for the Oakland Assembly question, and "Pennsylvania" (0.479) with abstraction markers "Federal" (0.507) for the WCDL question. These patterns demonstrate that the gated concept stream stores distributed representations of entities (0.6--0.8 similarity) and their semantic associations (0.4--0.5 similarity).This preservation mechanism directly explains AGCLR's performance advantage. In the Manor Township example, AGCLR maintains the Pennsylvania$\to$William Penn binding needed for correct generation, while vanilla CoCoNuT hallucinates "Henry David Thoreau" after losing this entity relationship across passes. Similarly, the Oakland Assembly case shows temporal context preservation ("War"/"World" similarities prevent drift to "World War II"), and the WCDL example demonstrates multi-hop reasoning where both the base entity ("Pennsylvania") and abstraction markers ("Federal"/"country") enable geographic generalization. These findings directly explain the 71\% better information retention measured in Figure~\ref{fig:hidden_state_similarity} and AGCLR's +3.6\% EM improvement over vanilla CoCoNuT.

\section{Conclusion}

We identified the concept bottleneck---where intermediate states are lost across passes---and proposed AGCLR with a gated concept stream to resolve it. AGCLR outperforms vanilla CoCoNuT across all three datasets, with gains compounding at deeper stages. On ProsQA, AGCLR reaches 96\% accuracy while vanilla degrades to 92\%, directly demonstrating that persistent memory resolves reasoning depth limitations. Ablations show gains stem from persistent memory, not parameters: freezing write after pass 2 yields only $-0.8\%$ EM loss, confirming early capture suffices. Our evaluation uses single-seed GPT-2 124M on three benchmarks; while consistent improvements across diverse reasoning tasks suggest the approach generalizes, scalability to larger models and broader task distributions remains to be validated. Future work includes scaling to larger base models and multi-seed evaluation.

\bibliography{main}
\bibliographystyle{icml2026}
\newpage
\appendix

\section{Extended Ablation Studies}

\subsection{Persistent Memory vs Additional Parameters}

To validate that AGCLR's performance gains stem from \textit{persistent memory mechanisms} rather than simply additional parameters, we conduct an ablation study examining the role of dynamic writing across reasoning passes. We compare four configurations: (1) Vanilla CoCoNuT baseline, (2) AGCLR without write gate (read and forget only), (3) AGCLR with write gate frozen after pass 2, and (4) Full AGCLR with all gates active.

The key question is whether the write gate's value lies in \textit{early information capture} (passes 1--2) or \textit{continuous refinement} across all passes. If dynamic writing throughout all passes were critical, we would expect significant performance degradation when the write gate is frozen. Conversely, if early capture combined with persistent retrieval is sufficient, performance should remain largely intact.

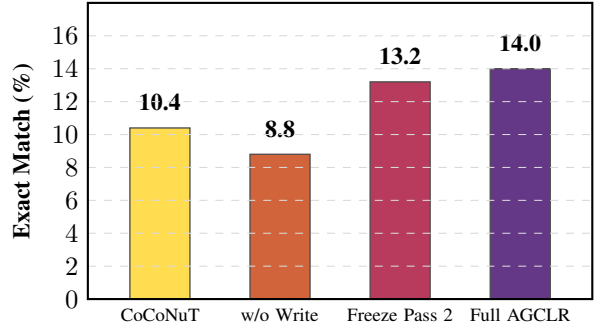
\begin{figure}[t]
\centering
\begin{tikzpicture}
\begin{axis}[
    width=\columnwidth,
    height=5.5cm,
    axis on top=true,
    ylabel={Exact Match (\%)},
    ylabel style={font=\bfseries\small},
    xtick={1,2,3,4},
    xticklabels={CoCoNuT, {w/o Write}, {Freeze Pass 2}, {Full AGCLR}},
    xticklabel style={font=\scriptsize, align=center, text width=1.6cm},
    xmin=0.4, xmax=4.6,
    ymin=0, ymax=18,
    ytick={0,2,4,6,8,10,12,14,16},
    ymajorgrids=true,
    grid style={dashed, gray!30},
    axis background/.style={fill=white},
    axis line style={line width=0.8pt, black},
    tick style={draw=none},
    axis lines*=box,
    ytick pos=left,
    xtick pos=bottom,
    clip=false,
]

\draw[fill={rgb,255:red,255;green,220;blue,80}, draw=black!70, line width=0.5pt]
    (axis cs:0.75,0) rectangle (axis cs:1.25,10.4);
\node[font=\small\bfseries, anchor=south] at (axis cs:1,10.4) [yshift=3pt] {10.4};

\draw[fill={rgb,255:red,210;green,100;blue,60}, draw=black!70, line width=0.5pt]
    (axis cs:1.75,0) rectangle (axis cs:2.25,8.8);
\node[font=\small\bfseries, anchor=south] at (axis cs:2,8.8) [yshift=3pt] {8.8};

\draw[fill={rgb,255:red,184;green,59;blue,94}, draw=black!70, line width=0.5pt]
    (axis cs:2.75,0) rectangle (axis cs:3.25,13.2);
\node[font=\small\bfseries, anchor=south] at (axis cs:3,13.2) [yshift=3pt] {13.2};

\draw[fill={rgb,255:red,100;green,55;blue,135}, draw=black!70, line width=0.5pt]
    (axis cs:3.75,0) rectangle (axis cs:4.25,14.0);
\node[font=\small\bfseries, anchor=south] at (axis cs:4,14.0) [yshift=3pt] {14.0};

\end{axis}
\end{tikzpicture}
\caption{\textbf{Ablation study of gating mechanisms.} Exact Match on HotpotQA.
\textit{CoCoNuT}: vanilla baseline (10.4\%).
\textit{w/o Write}: write gate removed (8.8\%).
\textit{Freeze Pass 2}: write gate frozen after pass~2 (13.2\%).
\textit{Full AGCLR}: all gates active (14.0\%).
The small gap between Freeze Pass~2 and Full AGCLR ($+0.8\%$ EM) shows early
capture suffices; the large drop without the write gate confirms it is critical.}
\label{fig:ablation_gating}
\end{figure}

Figure~\ref{fig:ablation_gating} presents our results. Remarkably, freezing the write gate after pass 2 results in only minimal performance degradation: 13.2\% EM versus 14.0\% EM for full AGCLR ($-0.8\%$ absolute). This near-equivalent performance demonstrates that early information capture is largely sufficient for multi-hop reasoning. Notably, the model without any write gate achieves only 8.8\% EM, confirming that the write gate is necessary---but its primary value lies in the initial passes. The read and forget gates then maintain and retrieve this early-captured information throughout subsequent reasoning steps.

This finding reveals AGCLR's core mechanism: the concept stream functions as \textit{persistent storage} rather than a dynamic scratchpad. Information is written once during early passes (1--2), then read and selectively forgotten across later passes (3--6). The write gate's primary value lies in identifying and capturing relevant information early, not in continuously refining the stored representation. This validates our hypothesis that AGCLR's gains emerge from the persistent memory architecture itself---the ability to store, maintain, and retrieve information across reasoning steps---rather than from simply adding more trainable parameters to the model.

\subsection{Individual Gate Ablations}

We systematically remove each gate by fixing its output to zero throughout training to understand their individual contributions. Results are presented in Table~\ref{tab:gate_ablation}.

\begin{table}[h]
\centering
\small
\begin{tabular}{lcc}
\toprule
\textbf{Method} & \textbf{HotpotQA EM (\%)} & \textbf{HotpotQA F1 (\%)} \\
\midrule
AGCLR (full)    & 14.0 & 19.4 \\
w/o write gate  & 8.8  & 17.1 \\
w/o read gate   & 9.4  & 17.8 \\
w/o forget gate & 8.4  & 18.9 \\
Vanilla CoCoNuT & 10.4 & 15.2 \\
\bottomrule
\end{tabular}
\caption{\textbf{Individual gate ablation results.} Each gate is removed by fixing its output to zero throughout training.}
\label{tab:gate_ablation}
\end{table}

Removing any single gate degrades performance, confirming all three components are necessary. The write gate is most structurally critical: without it, nothing is committed to the concept stream, dropping EM to 8.8\%, below vanilla CoCoNuT (10.4\%). However, the forget gate produces the \textit{largest performance drop} (8.4\% EM, $-5.6\%$), demonstrating that selective forgetting is essential for maintaining concept stream quality. In multi-hop reasoning, intermediate computations accumulate both relevant facts (e.g., entity names needed for later hops) and irrelevant context (e.g., formatting tokens, partial calculations from earlier steps). Without the forget gate pruning this noise, the concept stream becomes polluted across passes, degrading retrieval quality and preventing the model from isolating task-relevant information for final answer generation. Removing the read gate costs 4.6\% EM, confirming that cross-pass retrieval drives a meaningful share of AGCLR's gains. All three gates are indispensable to resolving the concept bottleneck.
\subsection{Initialization}

Gate weights $W_r, W_f, W_w$ are initialized to zero for stable warm-up. Gate biases use dataset-specific values (Table~\ref{tab:gate_init}): lower forget/higher write for ProsQA (preserves graph entities), higher forget/lower write for GSM8K (prunes intermediate steps).

\begin{table}[!ht]
\centering
\small
\begin{tabular}{lccc}
\toprule
\textbf{Dataset} & \textbf{Read} & \textbf{Forget} & \textbf{Write} \\
\midrule
GSM8K / HotpotQA & 0.43 & 0.27 & 0.18 \\
ProsQA           & 0.43 & 0.18 & 0.43 \\
\bottomrule
\end{tabular}
\caption{Dataset-adaptive gate initialization values $\sigma(b)$.}
\label{tab:gate_init}
\end{table}

This conservative initialization strategy prevents premature concept stream saturation. The low write gate activation (0.18) encourages the model to be selective about what information gets stored initially. The moderate forget gate activation (0.27) retains most information early in training, allowing the model to learn which facts are relevant. The read gate's balanced initialization (0.43) provides moderate retrieval strength. During training, gates adapt to dataset-specific patterns (Table~\ref{tab:gate_ablation}), demonstrating that the architecture learns task-appropriate memory management rather than relying on fixed heuristics.

\subsection{Concept Stream Content Analysis: Detailed Examples}

Table~\ref{tab:concept_stream_examples} provides detailed examples of what information gets written to the concept stream, showing cosine similarities between concept stream embeddings and vocabulary tokens for three contrastive cases where AGCLR answers correctly but vanilla CoCoNuT fails.

\begin{table*}[t]
\centering
\small
\begin{tabular}{p{4.5cm}p{2.5cm}p{4cm}p{3cm}}
\toprule
\textbf{Question} & \textbf{AGCLR Answer} & \textbf{Vanilla CoCoNuT Answer} & \textbf{Concept Stream Similarities} \\
\midrule
Who founded Manor Township, Pennsylvania? & William Penn & Henry David Thoreau & Penn: 0.806\\
& & & William: 0.681\\
& & & Pennsylvania: 0.614 \\
\midrule
When did Oakland Assembly close? & World War I & World War II & War: 0.760\\
& & & World: 0.684 \\
\midrule
What country is WCDL radio station in? & United States & Pennsylvania & Pennsylvania: 0.479\\
& & & Federal: 0.507\\
& & & country: 0.495 \\
\bottomrule
\end{tabular}
\caption{\textbf{Concept stream content examples.} Cosine similarities between concept stream embeddings and vocabulary tokens for three contrastive examples. High similarities (0.5--0.8) indicate successful storage of answer-relevant entities and their semantic associations.}
\label{tab:concept_stream_examples}
\end{table*}

These examples demonstrate three key patterns: (1) Entity binding preservation---AGCLR maintains the Pennsylvania$\to$William Penn relationship while vanilla CoCoNuT loses it and hallucinates a plausible but incorrect historical figure. (2) Temporal context retention---high similarity to "War" and "World" prevents drift to the more statistically common "World War II" after losing specific temporal markers. (3) Multi-hop reasoning---the concept stream preserves both base entities ("Pennsylvania") and abstraction markers ("Federal", "country"), enabling geographic generalization from state to country level, while vanilla CoCoNuT stops at the first hop.

\section{Implementation Details}

\subsection{Training Configuration}

All models were trained using AdamW optimizer with learning rate $5 \times 10^{-5}$, weight decay 0.01, and batch size 128 (8 per device $\times$ 16 gradient accumulation steps on A10 GPU). We used the curriculum learning schedule from \citet{hao2024coconut}, progressively increasing latent tokens from 0 to 6 across stages. Gate parameters were initialized following Meta's protocol: read gate bias at $-0.28$ ($\sigma^{-1}(0.43)$), forget gate at $-1.0$ ($\sigma^{-1}(0.27)$), and write gate at $-1.5$ ($\sigma^{-1}(0.18)$).

\subsection{Evaluation Details}

Exact Match (EM) and F1 scores were computed following standard QA evaluation protocols. For HotpotQA, we evaluated on 500 randomly sampled validation examples at stage 3 (6 latent tokens). For ProsQA, we report accuracy on the full validation set. All cosine similarity analyses used 100 validation samples at epoch 15, averaging across passes and examples.

\subsection{Computational Requirements}

AGCLR training on HotpotQA required approximately 15 epochs ($\sim$2 hours on a single NVIDIA A10 GPU). ProsQA training reached 96\% accuracy in 20 epochs on a single GH200 GPU. Total training time across all three datasets (GSM8K, HotpotQA, ProsQA) was approximately 8 hours on a single A10 GPU.

\section{Additional Limitations}

Beyond the limitations discussed in the main paper, we note several additional considerations:

\textbf{Single-seed evaluation.} All reported results are from single training runs. While consistent improvements across three diverse datasets suggest robustness, variance estimates from multiple seeds would strengthen claims.

\textbf{Model scale.} Our experiments use GPT-2 124M as the base model. Scalability to larger models (1B+ parameters) remains unexplored, though the architectural modifications are parameter-agnostic.

\textbf{Dataset coverage.} While we evaluate on arithmetic (GSM8K), multi-hop QA (HotpotQA), and planning (ProsQA), additional benchmarks (MuSiQue, 2WikiMultihopQA, StrategyQA) would further validate generalization.

\textbf{Comparison to retrieval systems.} We focus on controlled architectural comparisons against vanilla CoCoNuT. Comparison to retrieval-augmented systems or significantly larger models (e.g., GPT-3.5, GPT-4) is beyond our scope but would provide additional context.

\textbf{Interpretability depth.} Our concept-stream analysis follows a logit-lens-style probing; finer-grained feature attribution \cite{chaudhary2024evaluating, golechha2025modular} is left to future work.

\end{document}